# New wavelet-based superresolution algorithm for speckle reduction in SAR images

Mario Mastriani

**Abstract**—This paper describes a novel projection algorithm, the Projection Onto Span Algorithm (POSA) for wavelet-based superresolution and removing speckle (in wavelet domain) of unknown variance from Synthetic Aperture Radar (SAR) images. Although the POSA is good as a new superresolution algorithm for image enhancement, image metrology and biometric identification, here one will use it like a tool of despeckling, being the first time that an algorithm of super-resolution is used for despeckling of SAR images. Specifically, the speckled SAR image is decomposed into wavelet subbands, POSA is applied to the high subbands, and reconstruct a SAR image from the modified detail coefficients. Experimental results demonstrate that the new method compares favorably to several other despeckling methods on test SAR images.

*Keywords*—Projection, speckle, superresolution, synthetic aperture radar, thresholding, wavelets.

## I. INTRODUCTION

A SAR image is affected by speckle in its acquisition and processing. Image despeckling is used to remove the multiplicative speckle while retaining as much as possible the important signal features. In the recent years there has been an important amount of research on wavelet thresholding and threshold selection for SAR despeckling [1], [2], because wavelet provides an appropriate basis for separating noisy signal from the image signal. The motivation is that as the wavelet transform is good at energy compaction, the small coefficients are more likely due to noise and large coefficient due to important signal features [3]. These small coefficients can be thresholded without affecting the significant features of the image. Thresholding is a simple nonlinear technique, which operates on one wavelet coefficient at a time. In its basic form, each coefficient is thresholded by comparing against threshold, if the coefficient is smaller than threshold, set to zero; otherwise it is kept or modified. Replacing the small noisy coefficients by zero and inverse wavelet transform on the result may lead to reconstruction with the essential signal characteristics and with less noise.

Since the work of Donoho & Johnstone [3], there has been much research on finding thresholds, however few are specifically designed for images. Unfortunately, this technique has the following disadvantages:

1. it depends on the correct election of the type of thresholding (soft, hard, and semi-soft) or shrinkage, e.g., VisuShrink, SureShrink, OracleShrink, OracleThresh, NormalShrink, BayesShrink, Thresholding Neural Network (TNN), etc. [1]-[5],
2. it depends on the correct estimation of the threshold and the distributions of the signal and noise, which are unquestionably the most important design parameters of these techniques,
3. the specific distributions of the signal and noise may not be well matched at different scales.
4. it doesn't have a fine adjustment of the threshold after their calculation, and
5. it should be applied at each level of decomposition, needing several levels.

Therefore, a new method without these constraints will represent an upgrade. On the other hand, although considerable advances has been reported in superresolution [6]-[11], they have never been used as a denoising tool, and much less even like a despeckling tool of SAR images, at least, efficiently. Nevertheless, the superresolution algorithms are frequently used for image enhancement, image metrology and biometric identification, among others applications, where the noise is present.

## II. SPECKLE MODEL

Speckle noise in SAR images is usually modelled as a purely multiplicative noise process of the form

$$I_s(r,c) = I(r,c).S(r,c)$$
$$= I(r,c).[\,1+T(r,c)\,] \qquad (1)$$
$$= I(r,c) + N(r,c)$$

The true radiometric values of the image are represented by $I$, and the values measured by the radar instrument are represented by $I_s$. The speckle noise is represented by $S$. The parameters $r$ and $c$ means row and column of the respective pixel of the image. If $S'(r,c) = S(r,c) - 1$ and $N(r,c) = I(r,c)\,S'(r,c)$, one begins with a multiplicative speckle $S$ and finish with an additive speckle $N$ [12], which avoid the log-transform, because the mean of log-transformed speckle noise does not equal to zero [13] and thus requires correction to avoid extra distortion in the restored image.

For single-look SAR images, *S* is Rayleigh distributed (for amplitude images) or negative exponentially distributed (for intensity images) with a mean of 1. For multi-look SAR images with independent looks, *S* has a gamma distribution with a mean of 1. Further details on this noise model are given in [14].

III. PROJECTION ONTO SPAN ALGORITHM (POSA)

A. *POSA in wavelet domain as a despeckling tool (POSAshrink)*

One begins decomposing the speckled SAR image into four wavelet subbands [1]-[4]: Coefficients of Approximation (LL), and speckled coefficients of Horizontal Detail ($LH_s$), Vertical Detail ($HL_s$), and, Diagonal Detail ($HH_s$), respectively, as shown in Fig. 1, where: L means Low frequency, H means High frequency, DWT-2D is the Bidimensional Discrete Wavelet Transform, and IDWT-2D is the inverse of DWT-2D. The four wavelet subbands are orthogonal between them [6]. If an original image of *row-by-column* pixels is used, then each subbands will have *(row/2)-by-(column/2) pixels*.

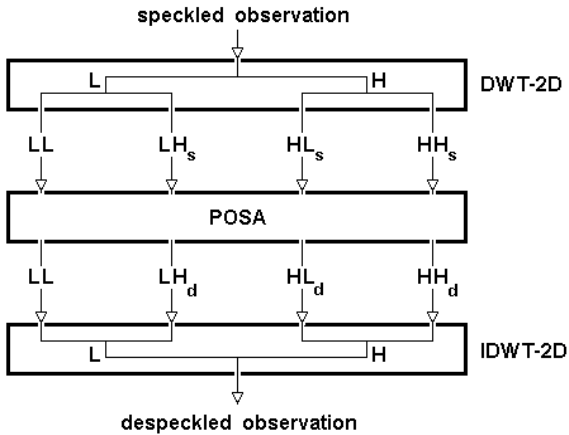

Fig. 1: POSA in wavelet domain as a despeckling tool.

Let {LL, $LH_s$, $HL_s$, $HH_s$} be a basis for an inner product space *W*. Let

$$\boldsymbol{LL} = LL / \|LL\|$$

$$\boldsymbol{LH_s} = LH_s / \|LH_s\| \qquad (2)$$

$$\boldsymbol{HL_s} = HL_s / \|HL_s\|$$

thus

$$LH_d = <LH_s, \boldsymbol{LL}> \boldsymbol{LL}$$

$$HL_d = <HL_s, \boldsymbol{LL}> \boldsymbol{LL} + <HL_s, \boldsymbol{LH_s}> \boldsymbol{LH_s} \qquad (3)$$

$$HH_d = <HH_s, \boldsymbol{LL}> \boldsymbol{LL} + <HH_s, \boldsymbol{LH_s}> \boldsymbol{LH_s} + <HH_s, \boldsymbol{HL_s}> \boldsymbol{HL_s}$$

where <A, B> means inner product of all real matrices A and B having the same number of columns [15], by <A, B> ≡ trace(A $B^T$). Finally, Equations (2) and (3) represent to the POSA. The reconstructed image (in this case the despeckling image) is the inverse of DWT-2D of the POSA output, as illustrated in Fig.1.

On the other hand, based on Eq.(1) POSAshrink does not need log-transform [12]. Besides, the most of times, the POSA is applied to the first level of decomposition exclusively, without the requirements of the thresholding method. Besides, the new method produced high quality, high-resolution image from a sequence of noisy, blurred and undersampled low-resolution frames. The frames are not restricted to being only displaced frame each other as in [11], [16], [17], but more general motion parameters between frames may be accommodated using the typical models [18], [19].

B. *POSA in wavelet domain as a superresolution tool*

Superresolution image reconstruction refers to the process of reconstructing a new image with a higher resolution using this collection of low resolution, shifted, rotated, and often noisy observations. This allows users to see image detail and structures which are difficult if not impossible to detect in the raw data. Superresolution is a useful technique in a variety of applications [20], and recently, researchers have begun to investigate the use of wavelets for superresolution image reconstruction [21]. A new method for superresolution image reconstruction based on the wavelet transform is necessary but in the presence of a very particular noise, the speckle [14].

*1) Typical superresolution algorithm based on wavelets*

The typical superresolution algorithm based on wavelets produces high-resolution (HR) image from a set of low-resolution (LR) frames. The relative motions in successive frames are estimated and used for aligning: HR image reconstruction from the set of LR images by performing image registration and then wavelet superresolution [6], [22]-[25].

The sample points in each frame into a HR grid. There are various types of models [18], [19] used to represent camera motion, namely, translation, rigid, affine, bilinear, and projective. The most general model is the projective model which has eight motion parameters. After registering all LR frames into a HR grid, the available samples distribute nonuniformly. This irregular sampling is called interlaced sampling. Then the wavelet superresolution algorithm will be applied in order to get the HR image.

*2) Proposed superresolution algorithm based on wavelets*

A new method for superresolution image reconstruction based on the wavelet transform is presented in the presence of speckle of unknown variance. To construct the superresolution image, an approach based on POSA is used.

*Case 1:* A Row x Column image is taken to be the original HR image. A (2 x 2) sensor array without sub-pixel displacement errors retrieves four Row/2 x Column/2 blurred and undersampled images (observations) {$O_1$, $O_2$, $O_3$, $O_4$}, which are corrupted by speckle, as shown in Fig.2.

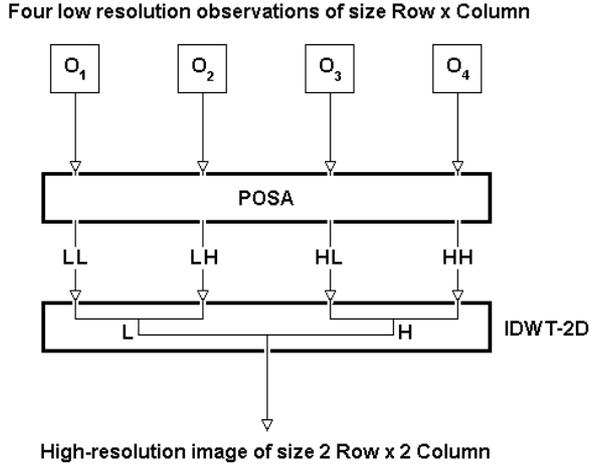

Fig. 2: POSA in wavelet domain as a superresolution tool (with four observations)

Let $\{O_1, O_2, O_3, O_4\}$ be a basis for an inner product space *W*. Let

$O_1 = O_1 / \|O_1\|$

$O_2 = O_2 / \|O_2\|$ (4)

$O_3 = O_3 / \|O_3\|$

thus

$LL = O_1$

$LH = <O_2, O_1> O_1$

$HL = <O_3, O_1> O_1 + <O_3, O_2> O_2$ (5)

$HH = <O_4, O_1> O_1 + <O_4, O_2> O_2 + <O_4, O_3> O_3$

*Case 2:* Now one has only one Row/2 x Column/2 blurred and undersampled image (observation), which are corrupted by speckle, then, three auxiliary matrices are used $\{A_1, A_2, A_3\} \in [0,1]$ of size Row/2 x Column/2 to feed POSA, as shown in Fig.3. Let $\{O, A_1, A_2, A_3\}$ be a basis for an inner product space *W*. Let

$O = O / \|O\|$

$A_1 = A_1 / \|A_1\|$ (6)

$A_2 = A_2 / \|A_2\|$

thus

$LL = O$

$LH = <A_1, O> O$

$HL = <A_2, O> O + <A_2, A_1> A_1$ (7)

$HH = <A_3, O> O + <A_3, A_1> A_1 + <A_3, A_2> A_2$

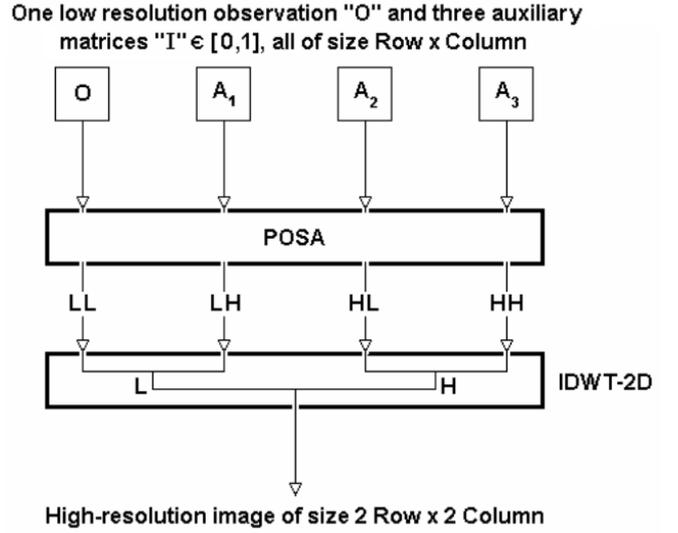

Fig. 3: POSA in wavelet domain as a superresolution tool (with one observation)

## IV. ASSESSMENT PARAMETERS FOR DESPECKLING AND EDGE PRESERVATION

In this work, the assessment parameters that are used to evaluate the performance of speckle reduction are Noise Variance, Mean Square Difference, Noise Mean Value, Noise Standard Deviation, Equivalent Number of Looks, Deflection Ratio, and Pratt's figure of Merit [26], [27].

### A. Noise Mean Value (NMV), Noise Variance (NV), and Noise Standard Deviation (NSD)

NV determines the contents of the speckle in the image. A lower variance gives a "cleaner" image as more speckle is reduced, although, it not necessarily depends on the intensity. The formulas for the NMV, NV and NSD calculation are

$$NMV = \frac{\sum_{r,c} I_d(r,c)}{R*C}$$

$$NV = \frac{\sum_{r,c}(I_d(r,c) - NMV)^2}{R*C} \quad (8)$$

$$NSD = \sqrt{NV}$$

where R-by-C pixels is the size of the despeckled image $I_d$. On the other hand, the estimated noise variance is used to determine the amount of smoothing needed for each case for all filters.

### B. Mean Square Difference (MSD)

*MSD* indicates average square difference of the pixels throughout the image between the original image (with speckle) $I_s$ and $I_d$, see Fig. 4. A lower *MSD* indicates a smaller difference between the original (with speckle) and despeckled image. This means that there is a significant filter performance. Nevertheless, it is necessary to be very careful with the edges. The formula for the *MSD* calculation is

$$MSD = \frac{\sum_{r,c}(I_s(r,c) - I_d(r,c))^2}{R*C} \quad (9)$$

### C. Equivalent Numbers of Looks (ENL)

Another good approach of estimating the speckle noise level in a SAR image is to measure the *ENL* over a uniform image region [1]. A larger value of *ENL* usually corresponds to a better quantitative performance. The value of *ENL* also depends on the size of the tested region, theoretically a larger region will produces a higher *ENL* value than over a smaller region but it also tradeoff the accuracy of the readings. Due to the difficulty in identifying uniform areas in the image, we proposed to divide the image into smaller areas of 25x25 pixels, obtain the *ENL* for each of these smaller areas and finally take the average of these *ENL* values. The formula for the *ENL* calculation is

$$ENL = \frac{NMV^2}{NSD^2} \quad (10)$$

The significance of obtaining both *MSD* and *ENL* measurements in this work is to analyze the performance of the filter on the overall region as well as in smaller uniform regions.

### D. Deflection Ratio (DR)

A fourth performance estimator used in this work is the *DR* proposed by H. Guo et al (1994), [2]. The formula for the deflection calculation is

$$DR = \frac{1}{R*C}\sum_{r,c}\left(\frac{I_d(r,c) - NMV}{NSD}\right) \quad (11)$$

The ratio *DR* should be higher at pixels with stronger reflector points and lower elsewhere. In H. Guo *et al*'s paper, this ratio is used to measure the performance between different wavelet shrinkage techniques. In this paper, the ratio approach to all techniques after despeckling in the same way [27] is applied.

### E. Pratt's figure of merit (FOM)

To compare edge preservation performances of different speckle reduction schemes, the Pratt's figure of merit is adopted [26] defined by

$$FOM = \frac{1}{\max\{\hat{N}, N_{ideal}\}}\sum_{i=1}^{\hat{N}}\frac{1}{1+d_i^2\alpha} \quad (12)$$

Where $\hat{N}$ and $N_{ideal}$ are the number of detected and ideal edge pixels, respectively, $d_i$ is the Euclidean distance between the *i*th detected edge pixel and the nearest ideal edge pixel, and $\alpha$ is a constant typically set to 1/9. *FOM* ranges between 0 and 1, with unity for ideal edge detection.

## V. EXPERIMENTAL RESULTS

### A. Despeckling

Here, a set of experimental results using one ERS SAR Precision Image (PRI) standard of Buenos Aires area is presented. For statistical filters employed along, i.e., Median, Lee, Kuan, Gamma-Map, Enhanced Lee, Frost, Enhanced Frost [1], [27], Wiener [5], DS [26] and Enhanced DS (EDS) [27], we use a homomorphic speckle reduction scheme [27], with 3-by-3, 5-by-5 and 7-by-7 kernel windows. Besides, for Lee, Enhan-ced Lee, Kuan, Gamma-Map, Frost and Enhanced Frost filters the damping factor is set to 1 [1], [27].

Fig. 4 shows a noisy image used in the experiment from remote sensing satellite ERS-2, with a 242-by-242 (pixels) by 65536 (gray levels); and the filtered images, processed by using VisuShrink (Hard-Thre-sholding), BayesShrink, NormalShrink, SUREShrink, and POSAShrink techniques respectively, see Table I.

All the wavelet-based techniques used Daubechies 1 wavelet basis and 1 level of decomposition (improvements were not noticed with other basis of wavelets) [4], [5], [26]. Besides, Fig. 4 summarizes the edge preservation performance of the POSAShrink technique vs. the rest of the shrinkage techniques with a considerably acceptable computational complexity.

Table I shows the assessment parameters vs. 19 filters for Fig. 4, where En-Lee means Enhanced Lee Filter, En-Frost means Enhanced Frost Filter, Non-log SWT means Non-logarithmic Stationary Wavelet Transform Shrinkage [12], Non-log DWT means Non-logarithmic DWT Shrinkage [13], VisuShrink (HT) means Hard-Thresholding, (ST) means Soft-Thresholding, and (SST) means Semi-ST [1]-[5].

The NMV and NSD are computed and compared over six different homogeneous regions in the choosed SAR image, before and after filtering, for all filters.

The POSAShrink has obtained the best mean preservation and variance reduction, as shown in Table I.

Since a successful speckle reducing filter will not signify-cantly affect the mean intensity within a homogeneous region, POSAShrink demonstrated to be the best in this sense too. The quantitative results of Table 1 show that the POSAShrink technique can eliminate speckle without distorting useful image information and without destroying the important image edges.

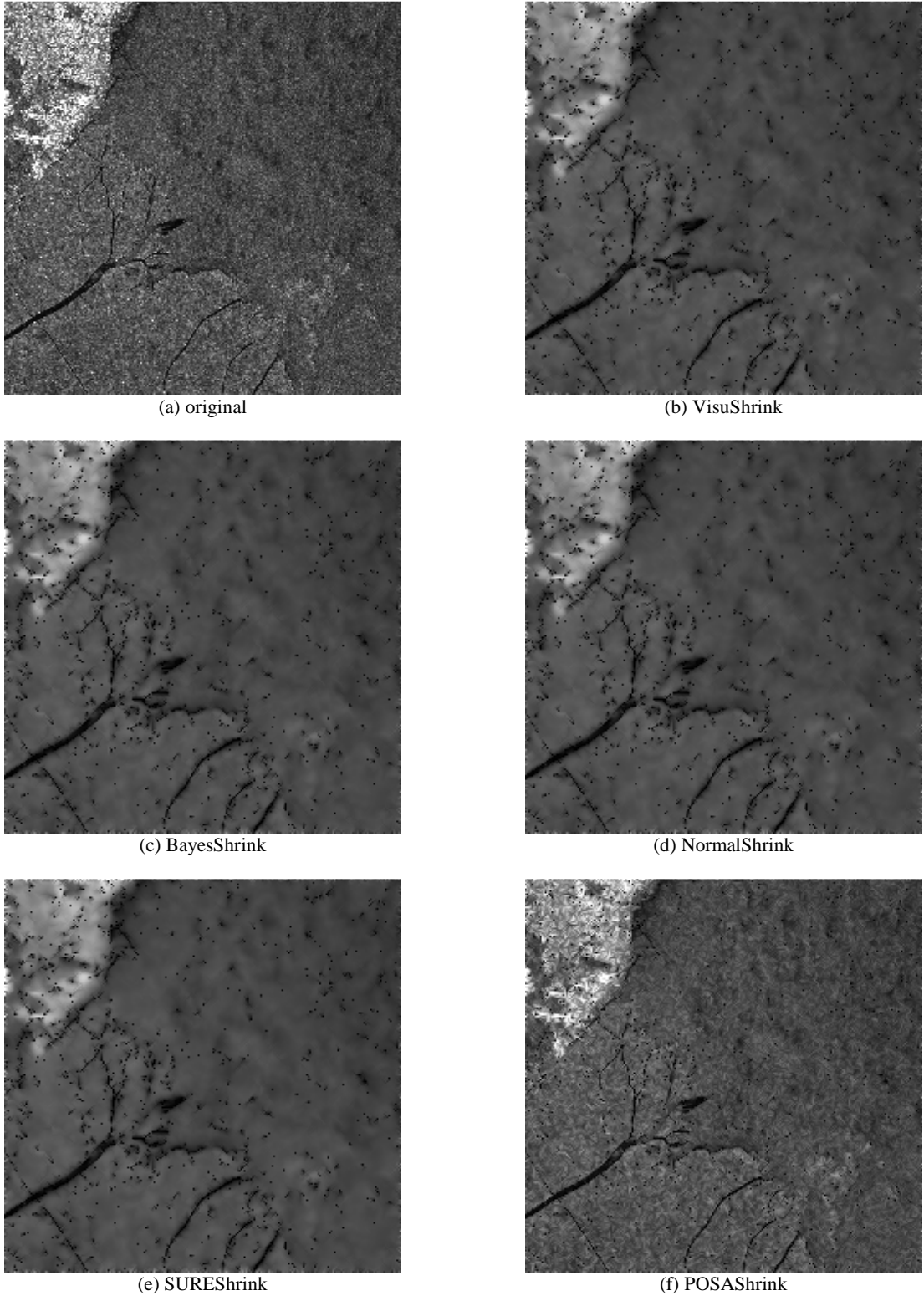

Fig. 4: Original and filtered images.

In fact, the POSAShrink outperformed the conventional and no conventional speckle reducing filters in terms of edge preservation measured by Pratt's figure of merit [26], as shown in Table 1. Fig.5 shows the histograms of the wavelet coefficients before shrinkage, after Visushrink (ST), after SUREshrink, and after POSAshrink.

TABLE I
ASSESSMENT PARAMETERS VS. FILTERS FOR FIG. 4.

| Filter | Assessment Parameters | | | | | |
|---|---|---|---|---|---|---|
| | MSD | NMV | NSD | ENL | DR | FOM |
| Original noisy image | - | 90.0890 | 43.9961 | 11.0934 | 2.5580e-017 | 0.3027 |
| En-Frost | 564.8346 | 87.3245 | 40.0094 | 16.3454 | 4.8543e-017 | 0.4213 |
| En-Lee | 532.0006 | 87.7465 | 40.4231 | 16.8675 | 4.4236e-017 | 0.4112 |
| Frost | 543.9347 | 87.6463 | 40.8645 | 16.5331 | 3.8645e-017 | 0.4213 |
| Lee | 585.8373 | 87.8474 | 40.7465 | 16.8465 | 3.8354e-017 | 0.4228 |
| Gamma-MAP | 532.9236 | 87.8444 | 40.6453 | 16.7346 | 3.9243e-017 | 0.4312 |
| Kuan | 542.7342 | 87.8221 | 40.8363 | 16.9623 | 3.2675e-017 | 0.4217 |
| Median | 614.7464 | 85.0890 | 42.5373 | 16.7464 | 2.5676e-017 | 0.4004 |
| Wiener | 564.8346 | 89.8475 | 40.3744 | 16.5252 | 3.2345e-017 | 0.4423 |
| DS | 564.8346 | 89.5353 | 40.0094 | 17.8378 | 8.5942e-017 | 0.4572 |
| EDS | 564.8346 | 89.3232 | 40.0094 | 17.4242 | 8.9868e-017 | 0.4573 |
| VisuShrink (HT) | 855.3030 | 88.4311 | 32.8688 | 39.0884 | 7.8610e-016 | 0.4519 |
| VisuShrink (ST) | 798.4422 | 88.7546 | 32.9812 | 38.9843 | 7.7354e-016 | 0.4522 |
| VisuShrink (SST) | 743.9543 | 88.4643 | 32.9991 | 37.9090 | 7.2653e-016 | 0.4521 |
| SureShrink | 716.6344 | 87.9920 | 32.8978 | 38.3025 | 2.4005e-015 | 0.4520 |
| NormalShrink | 732.2345 | 88.5233 | 33.3124 | 36.8464 | 6.7354e-016 | 0.4576 |
| BayesShrink | 724.0867 | 88.9992 | 36.8230 | 36.0987 | 1.0534e-015 | 0.4581 |
| Non-log SWT | 300.2841 | 86.3232 | 43.8271 | 11.2285 | 1.5783e-016 | 0.4577 |
| Non-log DWT | 341.3989 | 87.1112 | 39.4162 | 16.4850 | 1.0319e-015 | 0.4588 |
| POSAShrink | 867.1277 | 90.0890 | 32.6884 | 39.0884 | 3.2675e-015 | 0.4591 |

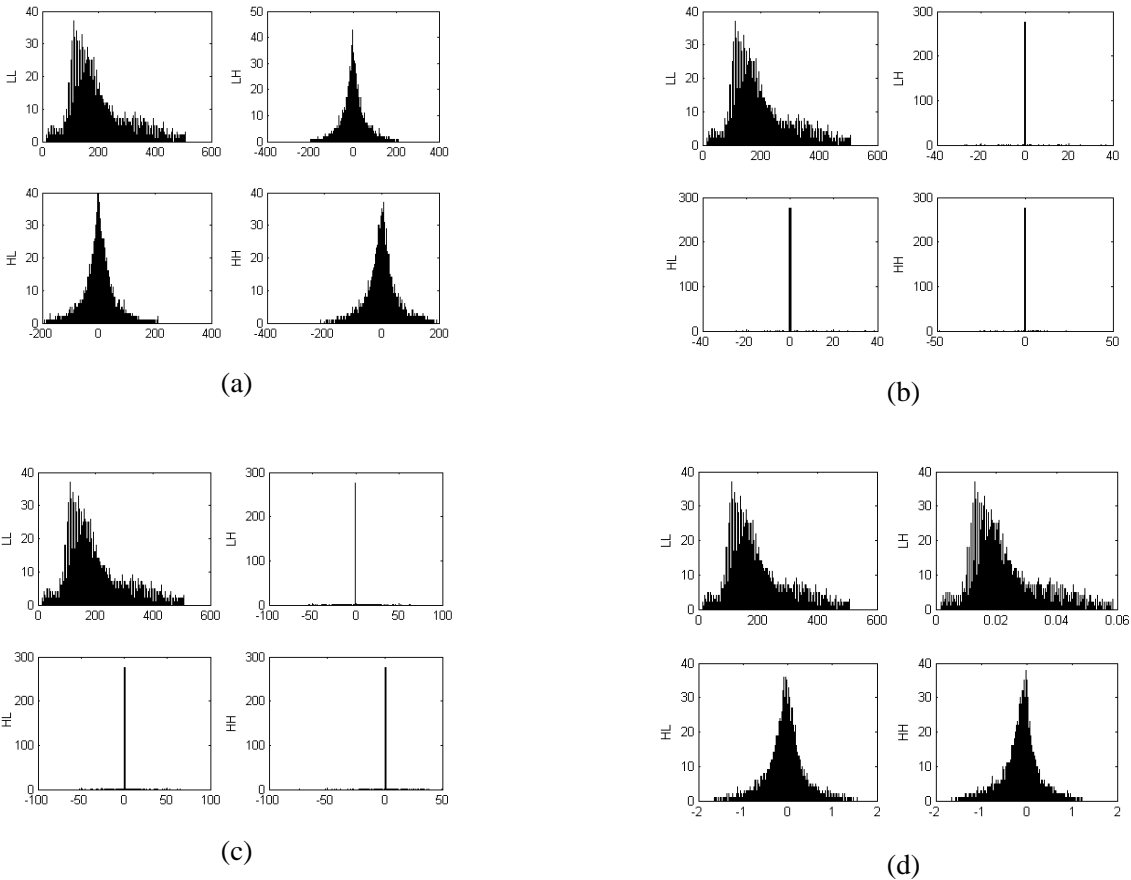

Fig. 5: Histograms of wavelet coefficients: (a) before shrinkage,
(b) after Visushrink (ST), (c) after SUREshrink, and (d) after POSAshrink.

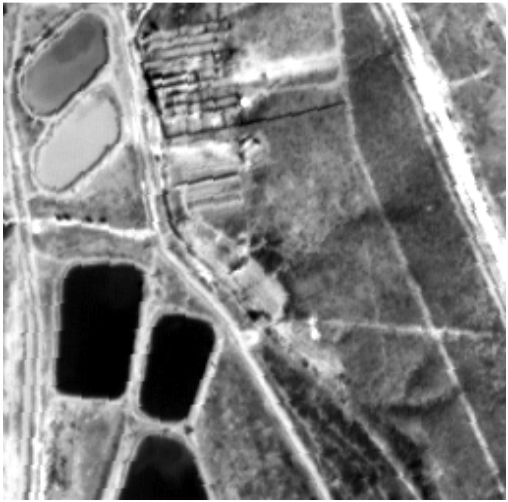
(a) Original image of size 256 x 256

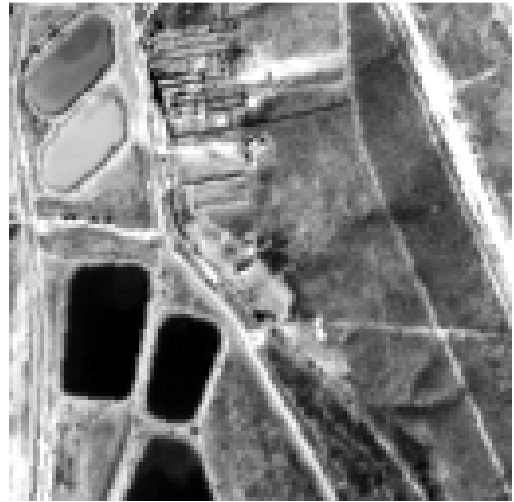
(b) One of four low resolution observations of size 128 x 128

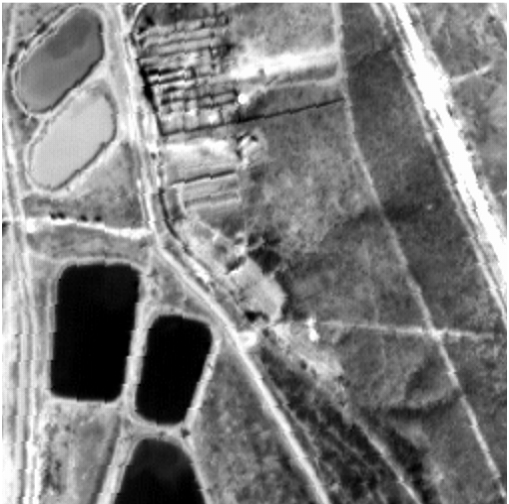
(c) High resolution image. PSNR = 26.4759 dB

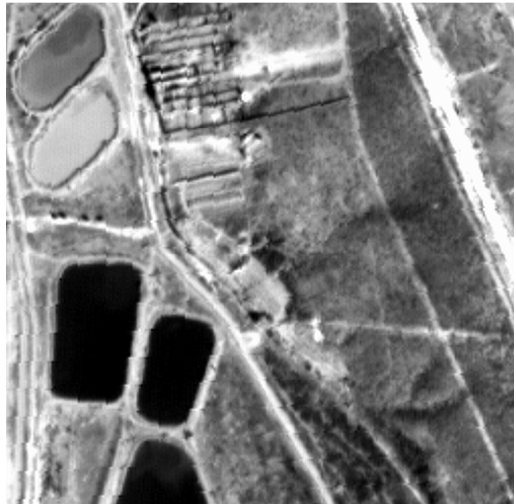
(d) High resolution image. PSNR = 26.6993 dB

Fig. 6: High-resolution image reconstruction.

*B. Superresolution*

Here, a 256 x 256 image (Fig.6(a)) is taken to be the original high-resolution image (with a resolution of 5 meters/pixel) of Sierra Grande, Patagonia. A (2 x 2) sensor array without sub-pixel displacement errors retrieves four 128 x 128 blurred and under sampled images, which are corrupted by speckle with a SNR of 30 dB. One of these low-resolution images is shown in Fig.6(b) and the image interpolated from these low resolution images is shown in Fig.6(c) with PSNR = 26.4759, and Fig.6(d) with PSNR = 26.6993.

In this experiment, we used Daubechies 4 wavelet basis and 1 level of decomposition.

On the other hand, both experiments were implemented in MATLAB® (Mathworks, Natick, MA) on a PC with an Athlon (2.4 GHz) processor.

## VI. EXPERIMENTAL RESULTS

A new speckle filter for SAR images based on wavelet denoising was presented. In order to convert the multiplicative speckle model into an additive noise model, Argenti *et al*'s approach is applied. The simulations show that the POSAShrink have better performance than the most commonly used filters for SAR imagery (for the studied benchmark parameters) which include statistical filters and several wavelets techniques in terms of smoothing uniform regions and preserving edges and features. The effectiveness of the technique encourages the possibility of using the approach in a number of ultrasound and radar applications. In fact, cleaner images suggest potential improvements for classification and recognition. Besides, considerably increased deflection ratio strongly indicates improvement in detection performance.

Finally, the method is computationally efficient and can significantly reduce the speckle while preserving the resolution of the original image, and avoiding several levels of decomposition and block effect.

On the other hand, the novelty of this paper is a new projection algorithm for superresolution for unknown blur. The POSA is a simple algorithm with a low computational complexity, where the blur not needs to be estimated. The novel has an excellent visual quality in presence of speckle. Such advantages were demonstrated in the simulations.

REFERENCES

[1] H.S. Tan. (2001, October). Denoising of Noise Speckle in Radar Image. [Online]. Available: http://innovexpo.itee.uq.edu.au/2001/projects/s804294/thesis.pdf
[2] H. Guo, J.E. Odegard, M. Lang, R.A. Gopinath, I. Selesnick, and C.S. Burrus, "Speckle reduction via wavelet shrinkage with application to SAR based ATD/R," Technical Report CML TR94-02, CML, Rice University, Houston, 1994.
[3] D.L. Donoho and I.M. Johnstone, "Adapting to unknown smoothness via wavelet shrinkage," *Journal of the American Statistical Association*, vol. 90, no. 432, pp. 1200-1224, 1995.
[4] S.G. Chang, B. Yu, and M. Vetterli, "Adaptive wavelet thresholding for image denoising and compression," *IEEE Transactions on Image Processing*, vol. 9, no. 9, pp.1532-1546, September 2000.
[5] X.-P. Zhang, "Thresholding Neural Network for Adaptive Noise reduction," *IEEE Transactions on Neural Networks*, vol.12, no. 3, pp.567-584, May 2001.
[6] N. K. Bose and S. Lertrattanapanich. Advances in wavelet superresolution. [Online]. Available: http://www.personal.psu.edu/users/s/x/sxl46/Bose01w.pdf
[7] S. P. Kim and N. K. Bose, "Reconstruction of 2-D bandlimited discrete signals from nonuniform samples," IEE Proceedings, Vol.137, No.3, Part F, June 1990, pp. 197-203.
[8] Seunghyeon Rhee and Moon Gi Kang, "Discrete cosine transform based regularized high-resolution image reconstruction algorithm," Optical Engineering, Vol.38, No.8, 1999, pp. 1348-1356.
[9] S. P. Kim, N. K. Bose and H. M. Valenzuela, "Recursive reconstruction of high resolution image from noisy undersampled multiframes," *IEEE Trans. on Acoust., Speech, and Signal Process.*, Vol.38, 1990, pp. 1013-1027.
[10] C. Ford and D. Etter, "Wavelet basis reconstruction of nonuniformly sampled data," *IEEE Trans. Circuits and System II*, Vol.45, No.8, August 1998, pp.1165-1168.
[11] N. Nguyen and P. Milanfar, "A wavelet-based interpolation-restoration method for supperresolution (wavelet superresolution)," Circuits Systems and Signal Process, Vol.19, No.4, 2000, pp.321-338.
[12] F. Argenti and L. Alparone, "Speckle removal from SAR images in the undecimated wavelet domain," *IEEE Trans. Geosci. Remote Sensing*, vol. 40, pp. 2363–2374, Nov. 2002.
[13] H. Xie, L. E. Pierce, and F. T. Ulaby, "Statistical properties of logarithmically transformed speckle," *IEEE Trans. Geosci. Remote Sensing*, vol. 40, pp. 721–727, Mar. 2002.
[14] J.W. Goodman, "Some fundamental properties of speckle," *Journal Optics Society of America*, 66:1145-1150, 1976.
[15] S.J. Leon, Linear Algebra with Applications, Maxwell Macmillan International Editions, New York, 1990.
[16] G.G. Walter, "Sampling Theorems and Wavelets," in Handbook of Statistics, Vol. 10, (eds. N.K. Bose and C.R. Rao), Elsevier Science Publishers B.V., North-Holland, Amsterdam, The Netherlands, 1993, pp. 883-903.
[17] M. Vetterli and J. Kovacevic, "Wavelets and Subband Coding," Prentice Hall PTR, Upper Saddle River, New Jersey 07458, 1995.
[18] S. Mann and R.W. Picard, "Video orbits of the projective group: A simple approach to featureless estimation of parameters," *IEEE Transactions on Image Processing*, Vol. 6, No.9, September 1997, pp.1281-1295.
[19] S. Lertrattanapanich and N.K. Bose, "Latest results on high-resolution reconstruction from video sequences," Technical Report of IEICE, DSP99-140, The Institution of Electronic, Information and Communication Engineers, Japan, December 1999, pp.59-65.
[20] R. Willett, R. Nowak, I. Jermyn, and J. Zerubia. Wavelet-Based Superresolution in Astronomy Astronomical Data Analysis Software and Systems XIII, ASP Conference Series, vol. 314, pp.107-116, 2004. [Online]. Available: http://www.adass.org/adass/proceedings/adass03/reprints/O2-1.pdf
[21] N.X. Nguyen. (2000, July). Numerical Algorithms for Image Superresolution. Ph.D. thesis, Stanford University. [Online]. Available: http://www.cse.ucsc.edu/~milanfar/NguyenPhDThesis.ps
[22] D.L. Ward. (2003, March). Redundant discrete wavelet transform based super-resolution using sub-pixel image registration. M.S. thesis, Department of the Air Force, Air University, Air Force Institute of Technology, Wright-Patterson Air Force Base, Ohio. [Online]. Available: https://research.maxwell.af.mil/papers/ay2003/afit/AFIT-GE-ENG--03-18.pdf
[23] S. Borman. (2004, April). Topics in multiframe superresolution restoration,"Ph.D. thesis, University of Notre Dame, Indiana. [Online]. Available: http://www.seanborman.com/publications/BormanPhD.pdf
[24] F.M. Candocia. (1998, May). A unified superresolution approach for optical and Synthetic Aperture Radar images. Ph.D. thesis, University of Florida. [Online]. Available: http://www.cnel.ufl.edu/bib/pdf_dissertation/candocia_dissertation.pdf
[25] F. Xiao, G. Wang, and Z. Xu, "Superresolution in two-color excitation fluorescence microscopy," *Optics Communications*, vol.228, pp. 225–230, 2003.
[26] Y. Yu, and S.T. Acton, "Speckle Reducing Anisotropic Diffusion," *IEEE Trans. on Image Process-ing*, vol. 11, no. 11, pp.1260-1270, 2002.
[27] M. Mastriani and A. Giraldez, "Enhanced Directional Smoothing Algorithm for Edge-Preserving Smoothing of Synthetic-Aperture Radar Images," *Journal of Measurement Science Review*, vol 4, no. 3, pp.1-11, 2004. [Online]. Available: http://www.measurement.sk/2004/S3/Mastriani.pdf